\documentclass[letterpaper, 10pt, conference]{ieeeconf}  
\IEEEoverridecommandlockouts
\usepackage{cite}
\usepackage{algorithm}
\usepackage{graphicx}
\usepackage{textcomp}
\usepackage{xcolor}
\usepackage{caption}
\usepackage{subcaption}
\usepackage{mathtools}
\usepackage{algpseudocode}
\usepackage{amsmath,amssymb,amsfonts}
\usepackage[font={small,it}]{caption}

\def\BibTeX{{\rm B\kern-.05em{\sc i\kern-.025em b}\kern-.08em T\kern-.1667em\lower.7ex\hbox{E}\kern-.125emX}}

\title{\LARGE \bf
Periodic and Event-Triggering for Joint Capacity Maximization and Safe Intersection Crossing}

\author{Christian Vitale, Panayiotis Kolios, and Georgios Ellinas
\thanks{The authors are with the Department of Electrical and Computer Engineering and the KIOS Research and Innovation Center of Excellence, University of Cyprus.
        {\tt\small \{vitale.christian, pkolios, gellinas\}@ucy.ac.cy}}
        \thanks{This work was supported in part by the European Union's Horizon 2020 Research and Innovation Programme under Grant 739551 (KIOS CoE - TEAMING) and Grant 101003439 (C-AVOID), and in part by the Government of the Republic of Cyprus through the Deputy Ministry of Research, Innovation and Digital Policy}}%

\begin{document}

\maketitle
\thispagestyle{empty}
\pagestyle{empty}

\begin{abstract}
Intersection crossing represents a bottleneck for transportation systems and Connected Autonomous Vehicles (CAVs) may be the groundbreaking solution to the problem. This work proposes a novel framework, i.e, \texttt{AVOID-PERIOD}, where an Intersection Manager (IM) controls CAVs approaching an intersection in order to maximize intersection capacity while minimizing the CAVs' gas consumption. Contrary to most of the works in the literature, the CAVs' location uncertainty is accounted for and periodic communication and re-optimization allows for the creation of safe trajectories for the CAVs. To improve scalability for high-traffic intersections, an event-triggering approach is also developed (\texttt{AVOID-EVENT}) that minimizes computational and communication complexity. \texttt{AVOID-EVENT} reduces the number of re-optimizations required by $92.2\%$, while retaining most of the benefits introduced by \texttt{AVOID-PERIOD}.
\end{abstract}

\section{INTRODUCTION}
Wireless communications have the potential to improve the performance of CAVs, especially in dangerous road sectors such as intersections, when on-board sensing is not sufficient to fully ascertain the status of an area-of-interest \cite{zhang2021emp}. Thus, CAVs have been the focus of several research works aiming towards a more efficient and safe management of intersection crossings.

A number of approaches have been investigated for CAV coordination while crossing an intersection: (i) reservation-based schemes, with intersections modeled as multi-agent problems and vehicles booking a specific area for intersection crossing \cite{levin2017conflict}; (ii) fuzzy controllers that navigate the CAVs at intersections \cite{rastelli2015fuzzy}; and (iii) optimization-based schemes based on Model Predictive Control, that aim at minimizing specific metrics, such as the time to cross the  intersection \cite{rios2016survey}. Nevertheless, the CAVs' present and future locations are generally considered error-free. Among the studies that consider location uncertainty, \cite{okamoto2018optimal} and \cite{chohan2019robust} present path planning algorithms able that respect pre-determined movement constraints, while \cite{nazari2018remote} also considers uncertainties due to unreliable wireless communication. Other approaches model the uncertainty of human-driven vehicles at intersections during trajectory planning \cite{hubmann2018automated}. While all previous works provided useful insights, when location uncertainty is accounted for, only our work in \cite{vitale2021itsc} considers a framework that optimizes the choices of CAVs aiming to maximize intersection capacity.

In this work, an IM collects CAVs' system status and associated uncertainties and decides the CAVs' future controls for crossing an intersection. In order to obtain safe trajectories, the IM decides controls following an optimal order for the CAVs, propagating in future time instants the location uncertainties of all vehicles for which controls have been already decided. Under this framework, and assuming a linear-Gaussian motion model, the IM is able to characterize, for each time instant, collision-free areas in the intersection that the CAV under optimization can safely use to move. Among all possible safe trajectories, a novel optimization (\texttt{AVOID-PERIOD}) is used to choose the optimal one, with the maximization of the intersection capacity as the primary objective and gas consumption as a secondary objective (i.e., when two or more equivalently optimal trajectories are available).

Contrary to our approach in \cite{vitale2021itsc}, \texttt{AVOID-PERIOD} exploits the communication between the IM and the CAVs; instead of optimizing the controls of the CAVs only once (at the entrance to the area supervised by the IM), it utilizes updated system state estimations transmitted by CAVs to periodically recompute their controls. Indeed, when compared with the corresponding forecasted system states, the ones transmitted by the CAVs exhibit less uncertainty and propagate a smaller error to future time instances, improving intersection capacity without compromising safety. Further, to improve scalability in high-density scenarios, an event triggering technique (\texttt{AVOID-EVENT}) is proposed to reduce the computational and communication complexity of \texttt{AVOID-PERIOD}. \texttt{AVOID-EVENT} identifies two main cases where triggering new controls for a CAV could be beneficial: (i) if an update from a possibly colliding CAV reduces the uncertainty associated with its future system state predictions, such that the center of the intersection can be crossed safely earlier than expected; (ii) if the distance traveled by a CAV is limited only by a preceding CAV on the same lane and an update allows to safely get closer to it. As a result, on average, \texttt{AVOID-EVENT} triggers a small percentage of the control updates triggered by \texttt{AVOID-PERIOD}, while retaining most of the performance gains exhibited by \texttt{AVOID-PERIOD}.

Summarizing, this work:
\begin{itemize}
\item builds on our uncertainty-aware mathematical framework in \cite{vitale2021itsc}, i.e., a periodic (the \texttt{AVOID-PERIOD}) control optimization which enables to exploit an updated and less error-prone view of the intersection;
\item develops an event-triggering-based optimization (i.e., \texttt{AVOID-EVENT})  which provides a trade-off between performance and computational/communication complexity;
\item presents an extensive simulation campaign to validate the proposed optimizations and showcases in detail their advantages and disadvantages.
\end{itemize}

\section{SYSTEM MODEL}
\label{sec:system_model}

\subsection{CAV's System State}
\label{subsec:systemmodel_estimation}
CAV $j$ follows discrete-time linear dynamics, with sampling interval $\delta \tau$, as shown below:
\begin{equation} \label{eq:agent_dynamics}
    x^j_\tau = \Phi x^j_{\tau-1} + \Gamma u^j_{\tau-1} + w^j_{\tau-1}
\end{equation}

\noindent where $x^j_\tau = [\text{x}^j,\dot{\text{x}}^j]_\tau^\top \in \mathbb{R}^4$ consists of position $\text{x}^j_\tau=[p_x, p_y]^j_\tau \in \mathbb{R}^2$ and velocity $\dot{\text{x}}^j_\tau = [\nu_x,\nu_y]^j_\tau \in \mathbb{R}^2$ components in 2D Cartesian coordinates at time $\tau$. To control each CAV,  acceleration controls are applied, with $u^j_\tau = [\text{a}^j_x, \text{a}^j_y]_{\tau}^\top \in \mathbb{R}^2$ denoting the applied acceleration vector at time $\tau$ and $w^j_\tau=[\text{w}^j,\dot{\text{w}}^j]_\tau^\top \in \mathbb{R}^4 \sim \mathcal{N}(0,\Sigma^j_w)$ denoting the Gaussian disturbance on the system because of uncontrolled forces on the CAV, that has zero mean and covariance matrix $\Sigma^j_w$. Being managed by the IM, with the chosen controls, it is assumed that CAVs only travel on north-to-south/south-to-north or west-to-east/east-to-west directions and do not change lanes while traversing the intersection. Thus, $\Gamma$ and $\Phi$ represent the unidimensional linear relationship between $x^j_\tau$ and pair $x^j_{\tau-1}$-$u^j_{\tau-1}$ \cite{vitale2021itsc}.
 
The dynamics of each CAV observe the Markov property (Eq. (\ref{eq:agent_dynamics})), i.e., a CAV's state at the next time step only depends on its current state and control input. Thus, to compute a vehicle's state $x_t, t\in [1,..,T]$ given a known initial state $x_0$ and a sequence of control inputs $u_{0:T-1}$ over the planning horizon $T$, Eq. (\ref{eq:agent_dynamics}) must be recursively applied as follows:
\begin{equation}\label{eq:CAV_traj}
    x_t = \Phi^t x_0 + \sum_{k=0}^{t-1} \Phi^k \left[ \Gamma u_{t-k-1} + w_{t-k-1} \right], \forall t
\end{equation}

\noindent where CAV index $j$ is not included in order to simplify the notation. Then, the trajectory of the CAV, $X_T = \{x_t\}, t \in [1,..T]$ over $T$, is a stochastic process, where each future state $x_t$ follows the distribution $x_t \sim \mathcal{N}(\mu_t,\Xi_t)$ having $\mu_t=[\boldsymbol{\mu},\dot{\boldsymbol{\mu}}]_t^\top$ and where $\Xi_t$ is given as follows:
\begin{equation}
\label{eq:state_prediction}
    \mu_t = \Phi^t x_0 + \sum_{k=0}^{t-1} \Phi^k \Gamma u_{t-k-1}, ~
    \Xi_t = \Sigma_0 +\sum_{k=0}^{t-1} \Phi^k \Sigma_w (\Phi^\top)^k.
\end{equation}
In Eq.~\ref{eq:state_prediction}, $\Sigma_0$ denotes the uncertainty associated with the initial system state of vehicle $j$. It should also be noted that can be \textit{easily pre-computed}, since $\Xi_t$ does not depend on the applied controls $u_{0:T-1}$. 

Finally, as in our work in \cite{vitale2021itsc}, with Gaussian measurement errors associated with the on-board sensors and GPS measurements through Kalman Filter (KF), CAV $j$ obtains at any time $\tau$ an estimation of its own system state which follows a multivariate Gaussian distribution with mean $\mu^\prime_\tau$ and covariance matrix $\Xi^\prime_\tau$. Such estimation is communicated to the IM and can be used as $x_0$ and $\Sigma_0$ in Eq. (\ref{eq:state_prediction}) (i.e., as the starting point of the control decisions).

\subsection{Uncertainty Characterization}
\label{subsec:uncertainty_estimation}
In order to account for potential collisions during path planning, the two-dimentional area that contains a CAV's barycenter at any time $\tau$ is modeled as an ellipse based on the CAV's location distribution, with probability $1\mathord{-}\epsilon$ ($\epsilon$ arbitrarily small). Depending on the IM's selected controls, if the ellipses that contain the two CAVs' barycenters never intersect, then their collision probability is bounded (i.e., there is a possibility of a collision only in the case where at least one of the two vehicles lies outside the ellipse). Thus, the maximum probability of collision, $P_c$, between vehicles $i$ and $j$ is given by:
\begin{equation}
\label{eq:bound_collision}
P_c \leq 1-(1-\epsilon)^2 \leq 2\epsilon-\epsilon^2\leq 2\epsilon
\end{equation}

Since $\Xi_t$ can be pre-computed $\forall t \in [1,..T]$, the size of the associated ellipse can be obtained by utilizing established statistical results, that consider the numerical integration of the distribution of the CAV's location \cite{ribeiro2004gaussian}. Specifically, with probability $1-\epsilon$ at time $t$, with $t\in [1,..T]$, the two semi-axes of the ellipse that contains the location of the CAV's barycenter are equal to:
\begin{equation}
\label{eq:ellipses_size}
\alpha^+_{t,j} = \sqrt{K_\epsilon\lambda^+_{t,j}};~~~\alpha^-_{t,j} = \sqrt{K_\epsilon\lambda^-_{t,j}};
\end{equation}

\noindent with $\lambda^+_{t,j}$ and $\lambda^-_{t,j}$ denoting the largest and smallest eigenvalues of the $\Xi_t$'s location sub-matrix and $K_\epsilon$ denoting the inverse of the cumulative density function of the chi-squared distribution having two degrees of freedom computed at $1\mathord{-}\epsilon$. 

\section{OPTIMIZATION FRAMEWORK}
\label{sec:optimization_framework}

\subsection{IM's Optimization Scenario Overview}
\label{subsec:IM_explanation}
\begin{figure}[thpb!]
      \centering
      \includegraphics[width=0.8\columnwidth]{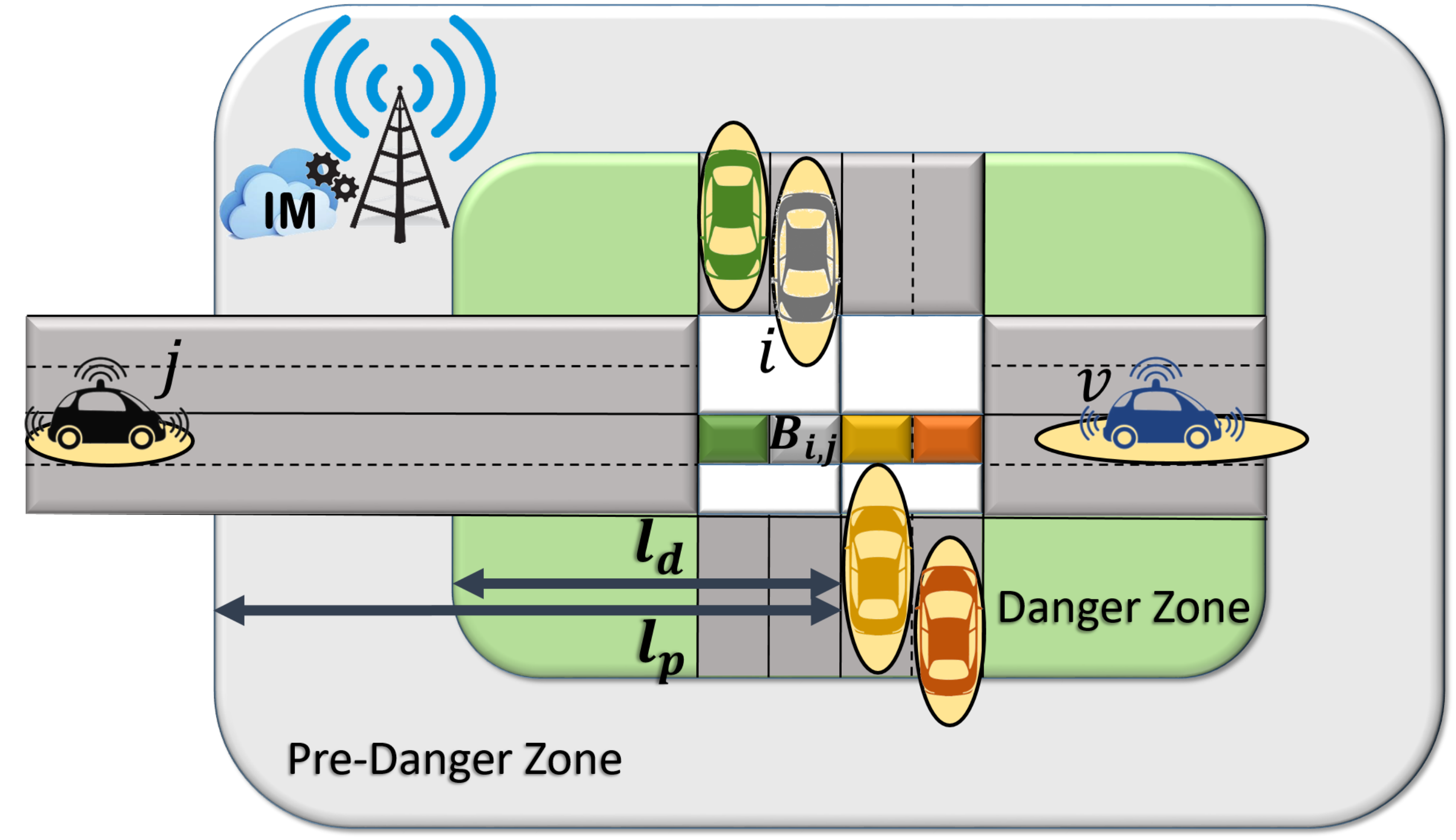}
      \caption{Overview of the IM's optimization scenario.}
      \label{fig:scenario}
   \end{figure}

In order to coordinate CAVs traversing an intersection, an IM is utilized. The area around the intersection is divided into the \textit{pre-danger zone} at a distance of $l_p$ meters from the intersection's center and the \textit{danger zone}, at a distance of $l_d$ m from the center of the intersection, with $l_d\mathord{<}l_p$. Interaction with the IM is initiated when the vehicles enter the pre-danger zone. The system under consideration is shown in Fig. \ref{fig:scenario}.

The IM's objective is to maximize the intersection's achievable capacity, where capacity in this case refers to the average number of CAVs that are safely admitted into the intersection. In order to achieve this objective, the distance that is traveled by each CAV $j$ after entering the pre-danger zone must be maximized, over a planning horizon $T$ (i.e., the intersection traversal time by a CAV must be minimized). With \texttt{AVOID-PERIOD}, to maximize the intersection capacity, the IM applies a receding horizon approach. When CAV $j$ enters the pre-danger zone, the IM computes controls $u^j_{0:T-1}$ over $T$, and communicates to $j$ the control $u^j_0$ for the next time slot. Subsequently, once the IM receives the CAV's updated system state estimation, obtained after applying the previous control decision, it computes the new control profile, i.e., $u^j_{1:T-1}$, over a planning horizon of length $T-1$, and transmits $u^j_1$ to CAV $j$. This operation is repeated until the end of the original planning horizon (i.e., $T$ times).

\subsection{\texttt{AVOID-PERIOD} Optimization}
\label{subsec:predanger_zone}
The \texttt{AVOID-PERIOD} technique jointly tackles both the danger zone demand management and the intersection capacity maximization at each time slot, in a receding horizon fashion, thus, extending the capabilities of a state-of-the-art intersection management algorithm, i.e., \texttt{AVOID-DM}, that we previously presented in \cite{vitale2021itsc} where joint danger zone demand management and intersection capacity maximization takes place only when the CAVs enter the pre-danger zone. 

In \texttt{AVOID-PERIOD}, CAVs' trajectories are computed sequentially, and the expected trajectories' information can be exploited to obtain safe trajectories for all the ones that follow in the planning order. Since the solutions chosen by the IM impose that CAVs do not change lanes while passing through the intersection, two collisions categories are studied: lateral collisions when vehicles cross the intersection following directions perpendicular to each other, and frontal collisions, when vehicles follow one another. In the case of lateral collisions for CAV $j$, a collision area $\mathcal{B}_{i,j}$ can be defined for each lane that crosses its path (see example in  Figure \ref{fig:scenario}). Subsequently, the probability of lateral collision bound in Eq.(\ref{eq:bound_collision}) is respected in the case where the ellipses that depict the expected barycenter locations of two CAVs with intersecting trajectories are never in the corresponding collision area at the same time. Similarly, the probability of frontal collision bound in Eq.(\ref{eq:bound_collision}) is respected in the case where the distance between CAV $j$'s ellipse and the preceding CAV $v$'s ellipse is at least $d_{min}=f_v+f_j+s$ (where $f_j$ denotes the distance between the barycenter of vehicle $j$ and its front bumper, $f_v$ denotes the distance between the barycenter of vehicle $v$ and its back bumper, and $s$ is the pre-defined safety distance).

To plan safe trajectories, the order used by the IM for deciding the CAVs' controls, at any time $\tau$, is critical. In this work, the planning order chosen at each time $\tau$ respects the order in which CAVs cross the center of the intersection. Thus, among two CAVs that may collide laterally, for the one passing first from the collision area, the IM does not need to know the trajectory of the other, while, for the one passing second, the IM needs to know the interval of time the collision area is occupied, in order to avoid ellipses intersecting one another. If the planning order respects the crossing order of CAVs from the center of the intersection, and considering that CAVs do not change lanes, CAVs following one another can also choose a safe trajectory. Indeed, at the moment of planning, the IM knows the expected locations, for any future time slot, of the preceding CAV's ellipse, and it can easily impose a constraint so to avoid frontal collisions.

Summarizing, at each time slot $\tau$, the IM determines the crossing order $\mathcal{O}_\tau$ of CAVs through the center of the intersection, based on the decision taken at $\tau\mathord{-}1$. For this purpose, the mean expected location of the CAVs' barycenter is used ($\mu$ in Eq. (\ref{eq:state_prediction})). Then, the IM plans a safe trajectory for each CAV in the planning order, with the objective of maximizing intersection capacity. Indexing CAVs in $\mathcal{O}_\tau$ as the order used for planning decisions, the IM decides the trajectory for $j$ through its acceleration profile $u^j_{\omega-1:T-1}$, with time $\tau$ corresponding to the $\omega\mathord{-}th$ time slot in the planning horizon of $j$, and ensures that the ellipse of $j$ does not intersect with any ellipse of CAVs $\{1,...,j-1\}$ in any time slot in $T$. 

The size of the ellipses (Eqs. \ref{eq:state_prediction} and \ref{eq:ellipses_size}) characterizes the uncertainty around future CAVs' system states. It should be noted that while the CAV is within the pre-danger zone, it exploits on-board sensors to avoid the only possible collision, i.e., with the preceding CAV. Thus, the pre-danger area is treated as any other road section that precedes the intersection and the IM considers that the uncertainty around the mean expected location $\mu$ follows a multivariate Gaussian distribution with zero-mean  and constant covariance matrix equal to $\Sigma_0+\Sigma^j_w$, with $\Sigma_0$ as the worst-case covariance matrix of the one-step Kalman Filter system state error estimation. On the contrary, within the danger zone, sensors on-board  the vehicles cannot always be used to avoid possible sudden lateral collisions. Thus, upon the CAV's entry into the danger zone, the IM propagates the error in future time slots as in Eq. (\ref{eq:state_prediction}). The reader should note that, even though the moments when CAVs $\{1,...,j-1\}$ enter the danger zone are known, this is not the case for the CAV $j$ under optimization prior to planning. Therefore, in \texttt{AVOID-PERIOD} the computation of the optimal time for vehicle $j$ to enter the danger zone is embedded in the optimization, effectively managing the demand at the entrance to the danger zone. To accomplish this, a binary variable $b_t$ is defined for each time $t\in [\omega,...,T]$ in the planning horizon of CAV $j$, that takes the value $0$ if vehicle $j$ is in the pre-danger zone, and the value $1$ otherwise. Subsequently, for computing the correct ellipse's semi-major axes at any time $t$ the following equation suffices:
\begin{equation}
\label{eq:axes_computation}
\alpha_j(t)= \alpha_{0,j}^+ - \sum_{k=\omega}^{t} b_k (\alpha_{t-k,j}^+-\alpha_{t-k+1,j}^+).
\end{equation}

Since a CAV traverses the danger zone upon entry, without returning to the pre-danger zone, the different terms in Eq. (\ref{eq:axes_computation}) cancel each other out, with the only remaining term corresponding to the number of time-slots during which the CAV is in the danger zone at time $t\in[\omega,...,T]$. Importantly, Eq. (\ref{eq:axes_computation}) holds even in the case when vehicle $j$ is already in the danger zone and all auxiliary binary variables equal $1$.

Based on the discussion above, \texttt{AVOID-PERIOD} is presented, that considers the $\omega\mathord{-}th$ time slot in the planning horizon of CAV $j$, which, w.l.o.g. traverses the intersection west-to-east. If not otherwise specified, $i\in\{1,...,j-1\}$, $t\in\{\omega,...,T\}$, and CAV $v$ precedes $j$ in the same lane.

\begin{subequations}
\small
\begin{alignat}{3}
&\textbf{Problem } \texttt{AVOID-PERIOD}: & & \nonumber\\
&\!\!\!\!\!\!\!\!\!\max\limits_{a^j_{\omega-1:T-1}} ~\boldsymbol{\mu}^j_T \!-\!\!\sum_{t=\omega}^{T}\xi_t M + \gamma\!\sum_{t=\omega}^{T} \boldsymbol{\mu}^j_t - \beta\!\!\!\!\sum_{t=\omega+1}^{T}\!|a^j_t - a^j_{t-1}| & &\label{eq:obj_avoid_dm}
\end{alignat}
\vspace{-0.4cm}
\begin{alignat}{3}
&\text{subject to:} &  & \nonumber\\
& \mu_t^i,~ \mu_t^j \text{ as in (\ref{eq:state_prediction})}& & \label{eq:eq1_avoid_dm}\\
& a^j_t \in [a_{MIN}, a_{MAX}], ~\boldsymbol{\dot{\mu}}^j_t \in [v_{MIN},v_{MAX}] & & \label{eq:eq4_avoid_dm}\\
& |a^j_t\mathord{-}a^j_{t\mathord{-}1}|\leq\Delta a & & \label{eq:eq6_avoid_dm}\\
& \alpha_j(t)  \text{ as in (\ref{eq:axes_computation})} && \label{eq:eq6b_avoid_dm}\\
& \boldsymbol{\mu}^j_t<-l_D+b_t M & &  \label{eq:eq6c_avoid_dm}\\
& b_t \in\{0,1\} & &  \label{eq:eq6d_avoid_dm}\\
& \boldsymbol{\mu}_t^v - \boldsymbol{\mu}_t^j \geq \alpha_j(t)+\alpha_v(t)+d_{min}-\xi_t & & \label{eq:eq7_avoid_dm}\\
& \xi_t\geq 0 & &  \label{eq:eq7b_avoid_dm}\\
& \boldsymbol{\mu}_t^j - \alpha_j(t)\mathord{\geq}\mathcal{B}_{i,j} ~||~\boldsymbol{\mu}_t^j + \alpha_j(t)\mathord{\leq}\mathcal{B}_{i,j} & &\nonumber\\
& \quad\quad\quad\quad\quad\quad\quad \rlap{$\displaystyle \forall t ~|~ \boldsymbol{\mu}_t^i \mathord{\pm}(\alpha_i(t) \mathord{+}f_i\mathord{+}s) \mathord{\in} \mathcal{B}_{i,j}$} & & \label{eq:eq8_avoid_dm}
\end{alignat}
\end{subequations}
\normalsize

\noindent where the $x$ direction subscript is not included for notational simplicity. To cope with the adopted receding horizon approach, slack variables $\xi_t$ are introduced to always find a solution and respect the presented safety constraints. As safety is paramount, the slack variables are multiplied by a large constant $M$ in the objective function (Eq. \ref{eq:obj_avoid_dm}). If slack variables are necessary, a solution is found that minimizes the slack variables' summation, i.e., the amount of violation of the safety constraints, practically ignoring any other component. Otherwise, $\xi_t=0$ $\forall t\in\{\omega,...,T\}$.

A multi-objective function is utilized when all cases where slack variables are not needed. The maximization of the distance traveled by vehicle $j$ is realized by maximizing CAV $j$'s $x$-axis mean location at the end of the optimization window $T$ (Eq. \ref{eq:obj_avoid_dm}). Further, the following fact is considered: if vehicle $i$ that crosses the intersection constraints the movement of vehicle $j$, then vehicle $j$ will be unable to pass the corresponding collision area before vehicle $i$, even if vehicle $j$ accelerates to the maximum possible speed. For this reason, multiple optimal acceleration profiles allow $j$ to cross the corresponding collision area right after $i$. To chose amongst all these profiles, two additional terms are included in the optimization objective function. The first term is utilized to push vehicle $j$ as close as possible to the danger zone, starting from the moment it enters the pre-danger zone, in order to clear the way for new CAVs. This is done by summing-up the distance traveled by vehicle $j$ at each time slot, over the planning horizon $T$ (travelling earlier on larger distance increases such summation). The second term improves the smoothness of the controls by minimizing the difference between consecutive accelerations applied by vehicle $j$. It should be noted that, in our implementation, the absolute values that are summed in this term are transformed into a series of additional linear constraints \cite{shanno1971linear}; this is done without impairing the proposed approach's complexity.  

In terms of \texttt{AVOID-PERIOD}'s constraints, system state predictions respect Eq. (\ref{eq:state_prediction}), the acceleration and speed of $j$ respect valid bounds (Eq. (\ref{eq:eq4_avoid_dm})) and successive acceleration controls do not vary more than $\Delta a$ m/s$^2$ (Eq. (\ref{eq:eq6_avoid_dm})) in order to increase user comfort. Further, Eq. (\ref{eq:eq6b_avoid_dm}) determines the ellipse's size that contains vehicle $j$'s barycenter with fixed probability, due to the determination of $j$'s entry time in the danger zone, while in Eq. (\ref{eq:eq6c_avoid_dm}), if vehicle $j$ is present in the danger zone at $t$, binary variable $b_t$ (that multiplies large constant $M$) becomes $1$, in order to satisfy the constraint. On the other hand, if vehicle $j$ is outside the danger zone, the constraint is always satisfied and binary variable $b_t$ can take any value. Nevertheless, to reduce the size of the ellipse associated to $j$ and allow a larger traveled distance, $b_t$ turns automatically to $0$ in the pre-danger zone. Finally, to ensure the coordination amongst CAVs the last three constraints are utilized, with the $\alpha_i$ and $\alpha_v$ values known when the optimization initiates. Specifically, Eq. (\ref{eq:eq7_avoid_dm}) ensures that the distance between the barycenters of the ellipses of vehicle $j$ and its preceding vehicle $v$ is larger than $d_{MIN}$. If this constraint cannot be satisfied, in order to obtain a viable solution a slack variable $\xi_t\geq 0$ is used, allowing a small violation. Finally, by using Eq. (\ref{eq:eq8_avoid_dm}) vehicle $j$ traverses collision area $\mathcal{B}_{i,j}$, before or after $i$; i.e., when vehicle $i$ traverses $\mathcal{B}_{i,j}$,  vehicle $j$'s predicted location (plus (minus) the larger axes of its ellipse) is outside $\mathcal{B}_{i,j}$. 

\section{An Event Triggering Approach}
\label{subsec:tracking_opt}
An event-triggering technique (\texttt{AVOID-EVENT}) is further developed, aiming to retain the benefits of \texttt{AVOID-PERIOD}, while greatly reducing communication and computational overhead. Specifically, \texttt{AVOID-EVENT} triggers a new CAV optimization only in the pre-danger zone and only if it is strictly required. No optimization is triggered in the danger zone, since errors are propagated and a safe, albeit suboptimal, trajectory is always possible. In the pre-danger zone, in case the CAVs colliding with $j$ continue on their expected trajectories and with the same expected uncertainty, re-optimizing controls may be superfluous. Hence, $j$'s controls are re-optimized only if there exist changes between the intersection's expected future status at time $\tau$ and its expected status at the moment $j$'s trajectory was selected. 

For lateral collisions, the intersection occupancy metric, $I^j_\tau$, is introduced, representing at each time $\tau$, the last future time slot in which, before the crossing of $j$, any of the possible collision areas $\mathcal{B}_{i,j}$ are occupied by another CAV with a probability larger than $1\mathord{-}\epsilon$. Specifically,  
\begin{equation}
I^j_\tau = \!\! \max_{i\in\{1,...,j-1\}} \!\! \Big[ \max_{\tau^* | t^* \in [\omega^i-1:T-1]} \int_{\mathcal{B}_{i,j}} \!\!\!\!\!\!\! \mathcal{N}(\mu^i_{t^*},\Xi^i_{t^*}) ~dx dy \! > \! 1\mathord{-}\epsilon \Big]
\end{equation}

\noindent where $\tau^*$ corresponds to the $t^*$-th time slot in the planning horizon of CAV $i$, $\forall i\in\{1,...,j-1\}$ and where $I^j_\tau$ is computed exclusively on the set of CAVs crossing the intersection before $j$. Hence, for lateral collisions, a new optimization is triggered if $I^j_\tau<I^j$, with $I^j$ being the intersection occupancy at the moment the controls for CAV $j$ were last updated. In practice, a new optimization is triggered if the center of the intersection, from the perspective of CAV $j$, is free earlier compared to when its controls were last updated. If this is the case, CAV $j$ can: (i) anticipate its entrance in the center of the intersection, (ii) reduce its intersection traversal time, and (iii) ultimately increase the capacity of the intersection.

If $j$ is not impeded laterally, \texttt{AVOID-EVENT} may trigger a new optimization if there exist changes on the expected future system states of CAV $v$ that precedes $j$ (i.e., the location of $v$ is such that the distance that can be traveled by $j$ increases). This occurs when $d_T^v(\tau)>d_T^v$, where $d_T^v(\tau)$ is the point of $v$'s ellipse that is closer to the entrance to the pre-danger zone at the end of its planning horizon $T$, as computed at time $\tau$ and $d_T^v$ is the point of $v$'s ellipse that is closer to the entrance of the pre-danger zone at the end of its planning horizon $T$, as computed the last time the controls of $j$ were updated.

In general, the events triggered by \texttt{AVOID-EVENT} may occur in two main cases: (i) the CAVs possibly colliding with $j$ are able to accelerate, leaving more space to $j$; (ii) the uncertainty related to the future system states of the CAVs possibly colliding with $j$ reduces. Recalling that in the danger zone system state prediction errors are propagated to all future time slots, when a CAV updates its system state from the danger zone, errors are propagated for fewer time slots, hence reducing the associated uncertainty. In practice, with the events belonging to the aforementioned second case, any time that additional knowledge on colliding CAVs is available, \texttt{AVOID-EVENT} tries to exploit the possibility to increase the intersection capacity and triggers a new optimization for $j$.

If no new optimization is triggered, to cope with any deviation due to uncertainty within the pre-danger zone, $j$ tracks the last information (i.e., expected trajectory and speed) received from the IM to traverse safely the intersection area. Specifically, through the exploitation of a receding horizon approach and an updated system state estimation utilizing a KF, vehicle $j$ can minimize the error between the target system states that are computed through \texttt{AVOID-PERIOD} and all of vehicle $j$'s future expected system states following the technique described below:

\begin{subequations}
\small
\begin{alignat}{3}
&\textbf{Problem } \texttt{Car-Follow}: & & \nonumber\\
&\min\limits_{a^j_{w-1:T-1}} ~\sum_{t=\omega}^T |\boldsymbol{\mu}^j_t \mathord{-} \boldsymbol{m}^j_t| + \delta \tau \sum_{t=\omega}^T |\boldsymbol{\dot{\mu}}^j_t \mathord{-} \boldsymbol{\dot{m}}^j_t| \!+\!\!\sum_{t=\omega}^{T}\xi_t M& & &\label{eq:obj_avoid_car}\\
&\text{subject to:} &  & \nonumber\\
& \mu_t^v,~ \mu_t^j \text{ as in (\ref{eq:state_prediction})} & & \label{eq:eq1_avoid_car}\\
& a^j_t \in [a_{MIN}, a_{MAX}], ~\boldsymbol{\dot{\mu}}^j_t \in [v_{MIN},v_{MAX}] & &  \label{eq:eq6_avoid_car}\\
& \boldsymbol{\mu}_t^v - \boldsymbol{\mu}_t^j \geq \alpha_{0,j}^+\alpha_v(t)+d_{min}-\xi_t & & \label{eq:eq7_avoid_car}\\
& \xi_t\geq 0 & & \label{eq:eq8_avoid_car}
\end{alignat}
\end{subequations}
\normalsize

For minimizing the error to the target trajectory the objective function's (Eq. (\ref{eq:obj_avoid_car})) absolute values are modified into linear constraints, where the expected location and speed that have to be tracked are represented by $\boldsymbol{m}^j_t$ and $\boldsymbol{\dot{m}}^j_t$, respectively. To obtain the CAV's future system state predictions, the motion dynamics of the CAV are exploited (Eq. (\ref{eq:state_prediction})). Constraint~\eqref{eq:eq6_avoid_car} limits the values assumed by the controls and by the speed of the CAV, while Constraint \eqref{eq:eq7_avoid_car} is used to avoid frontal collisions, by forcing CAV $j$ to respect a minimum distance with its preceding CAV $v$. Again, a set of minimized slack variables $\xi_t$, $\forall t \in [\omega,...,T]$, is introduced to obtain a solution even when the uncertainty is such that Eq. (\ref{eq:eq7_avoid_car}) cannot be respected. A summary of \texttt{AVOID-EVENT} is given in Alg. \ref{algo:flow_info}.
\begin{algorithm}
\footnotesize
	\caption{The \texttt{AVOID-EVENT} approach.}\label{algo:flow_info}
	\begin{algorithmic}[1]
	\State $I^j$, $d_T^v$ initialized to $+\infty$, $-\infty$, resp., when $j$ enters pre-danger zone
		\For {$\forall$ time $\tau$}
			\For {$\forall j\in O_\tau$}
			\If {$j$ in pre-danger zone} IM computes $I_\tau^j$ and $d_T^v(\tau)$
				\If {$I_\tau^j<I^j$}
					\State IM computes $u^j_{\omega-1:T-1}$ w/ \texttt{AVOID-PERIOD}, sends to $j$
					\State $I^j=I_\tau^j$
				\ElsIf {$I^j=+\infty$ $\&$ $d_T^v(\tau)>d_T^v$}
					\State IM computes $u^j_{\omega-1:T-1}$ w/ \texttt{AVOID-PERIOD}, sends to $j$
					\State $d_T^v=d_T^v(\tau)$
				\Else ~$j$ computes $u^j_{\omega-1:T-1}$ w/ \texttt{CAR-Follow}
				\EndIf
			\EndIf
			\If {$j$ in danger zone} $j$ applies last received $u^j_{\omega-1:T-1}$
			\EndIf
			\EndFor
		\EndFor
	\end{algorithmic} 
\end{algorithm}

\normalsize

\section{PERFORMANCE EVALUATION}
\label{sec:performance_evaluation}

\subsection{Simulation Scenario}
\label{subsec:sim_scenario}
Figure \ref{fig:scenario} illustrates the 4-way intersection simulation scenario used for the performance comparions of \texttt{AVOID-PERIOD}, \texttt{AVOID-EVENT}, and \texttt{AVOID-DM} \cite{vitale2021itsc}, with the pre-danger and danger zones starting at $300$ m and at $150$ m from the center of the intersection, respectively. The optimization window chosen,  $T=56$ s,  allows for a CAV to traverse the entire danger zone when traveling with an average speed of $8$ m/s. Further, the sampling rate $\delta \tau=0.5$ s. The inter-arrival times for CAVs entering the danger zone follow the exponential distribution, with a mean equal to $2$ s, that corresponds to the average arrival rate of cars at a centrally-located intersection during peak hour within a medium-size city \cite{makrigiorgis2020extracting}. The CAVs' initial speed (between $0$ and $14$m/s) and their lane are selected using a uniform distribution. Finally, CAVs can enter the pre-danger zone only if at least one acceleration profile is feasible so that a collision with the preceding CAV can be avoided. 
   
Moreover, typical sensor sensitivity data \cite{chow2011evaluating} are utilized to model the uncertainty of the initial state, and of acceleration/GPS measurements used in the KF by the CAVs and in the system state prediction by the IM. Thus, (i) the worst-case approximation of the KF's covariance is $\Sigma_0=[0.6m^2, 0.2(m/s)^2; 0.2m^2, 0.06(m/s)^2]$, while (ii) the dynamics error's covariance is $\Sigma_w=[0.0125\delta \tau^4~0.025 \delta \tau^3;~0.025 \delta \tau^3~0.5 \delta \tau^2]$. For both cases, the covariance matrix holds for location and speed in the direction of the CAV's movement, and it is zero otherwise. Same-size CAVs are assumed, having a distance between their barycenters of $d_{min}=8$ m in order to ensure a safe distance of at least $4$ m. Also, the acceleration takes (absolute) values in the range $0$ to $3$m/s$^2$, while $\Delta a= 1$m/s$^2$. Finally, $\epsilon=10^{-5}$, $\gamma=10^{-6}$, $\beta=10^{-5}$. Each simulation involves $10^4$ CAVs.

The reader should note that, \texttt{AVOID-DM}, that selects the controls of the CAVs (under location uncertainty) only upon arrival to the pre-danger zone, without further exploiting the updated system state estimations communicated by the CAVs to the IM, is used in these simulations as a benchmark.  


 \begin{figure}[thpb!]
      \centering
      \includegraphics[width=0.8\columnwidth]{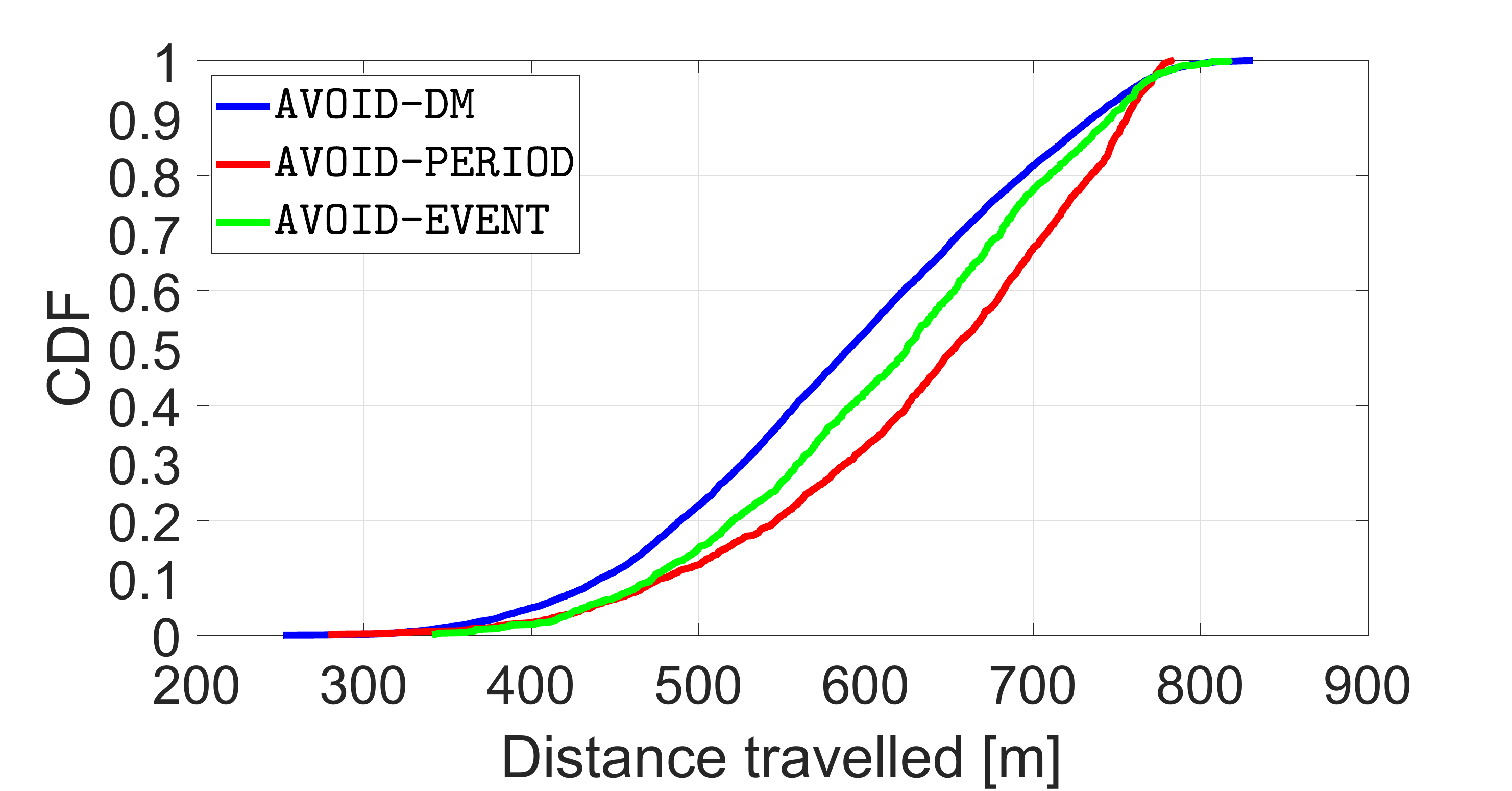}
      \caption{CAVs' traveled distance in $T$.}
      \label{fig:distance_travelled}
   \end{figure}

\subsection{Performance of \texttt{AVOID-PERIOD} and \texttt{AVOID-EVENT}}
\label{subsec:perf_AVOID}
The performance of the presented approaches was evaluated utilizing a  MATLAB simulator that was developed to model at each time slot the predicted, estimated, and real position of CAVs in the intersection, while all optimizations were solved using GUROBI.

Fig. \ref{fig:distance_travelled} shows the cumulative distribution function of the total distance the CAVs have traveled in their planning horizon. Exploiting the updated system state and uncertainty prediction communicated by CAVs to the IM, \texttt{AVOID-PERIOD} improves the distance traveled by CAVs, compared to the state-of-the-art \texttt{AVOID-DM}, reaching a gain of $12.26\%$ in the tail of the distribution, i.e., where it counts the most.  Similarly, \texttt{AVOID-EVENT} also improves the distance traveled by vehicles with gains up to $11.15\%$ in the first percentiles of the CDF. Such performance improvement is critical, especially in high-density scenarios, with both \texttt{AVOID-EVENT} and \texttt{AVOID-PERIOD} exploiting additional information during planning by the IM. Indeed, accounting for a more accurate system state estimation and prediction when compared with the one available at the entrance to the pre-danger zone, allows originating trajectories that result in CAVs traveling longer distances. 

Also, in \texttt{AVOID-EVENT}, on average, for each CAV, the number of events triggering a new control update equals $8.786$, contrary to \texttt{AVOID-PERIOD} that triggers a new control each time slot, i.e., $112$ times for each CAV. Thus, while both techniques exhibit similar performance, especially at the tail of the traveled distance CDF, the computational complexity of \texttt{AVOID-EVENT} is reduced by $92.2\%$. Additionally, as the IM sends control updates only when a new event is triggered, the CAVs' downlink traffic is also reduced by $92.2\%$ (no effect on uplink traffic as the CAVs send an update on their system state prediction at every slot).
 \begin{figure}[thpb!]
      \centering
      \includegraphics[width=0.8\columnwidth]{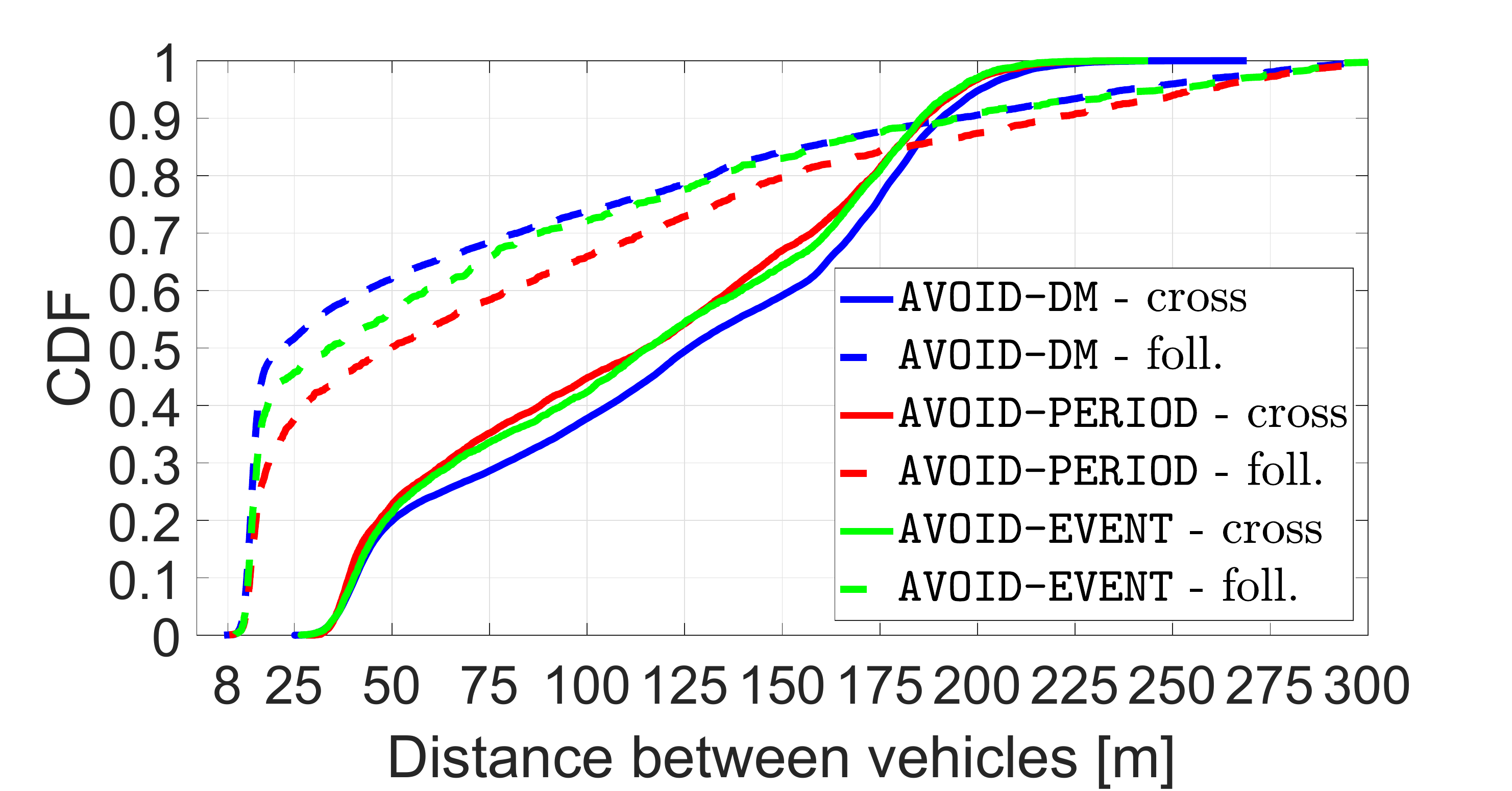}
      \caption{Minimum distance between CAVs that share a possible collision area.}
      \label{fig:distance_vehicles}
   \end{figure}   

All CAVs, for all approaches considered, respect the minimum distance of $8$ m between CAVs that allows avoiding collisions (Fig. \ref{fig:distance_vehicles}). Further, this distance is smaller for CAVs that follow each other, as compared to CAVs that cross each other's trajectory. This is the case, as the presented optimizations exploit a conservative approach for CAVs in the center of the intersection, allowing at most one ellipse in each of the collision areas. Further, as expected, \texttt{AVOID-PERIOD} and \texttt{AVOID-EVENT} perform better than \texttt{AVOID-DM}, since the additional knowledge exploited at the moment of planning, allows CAVs to safely get closer to each other. 

   \begin{figure}[thpb!]
\centering
\begin{subfigure}{.5\textwidth}
  \centering
  \includegraphics[width=0.8\columnwidth]{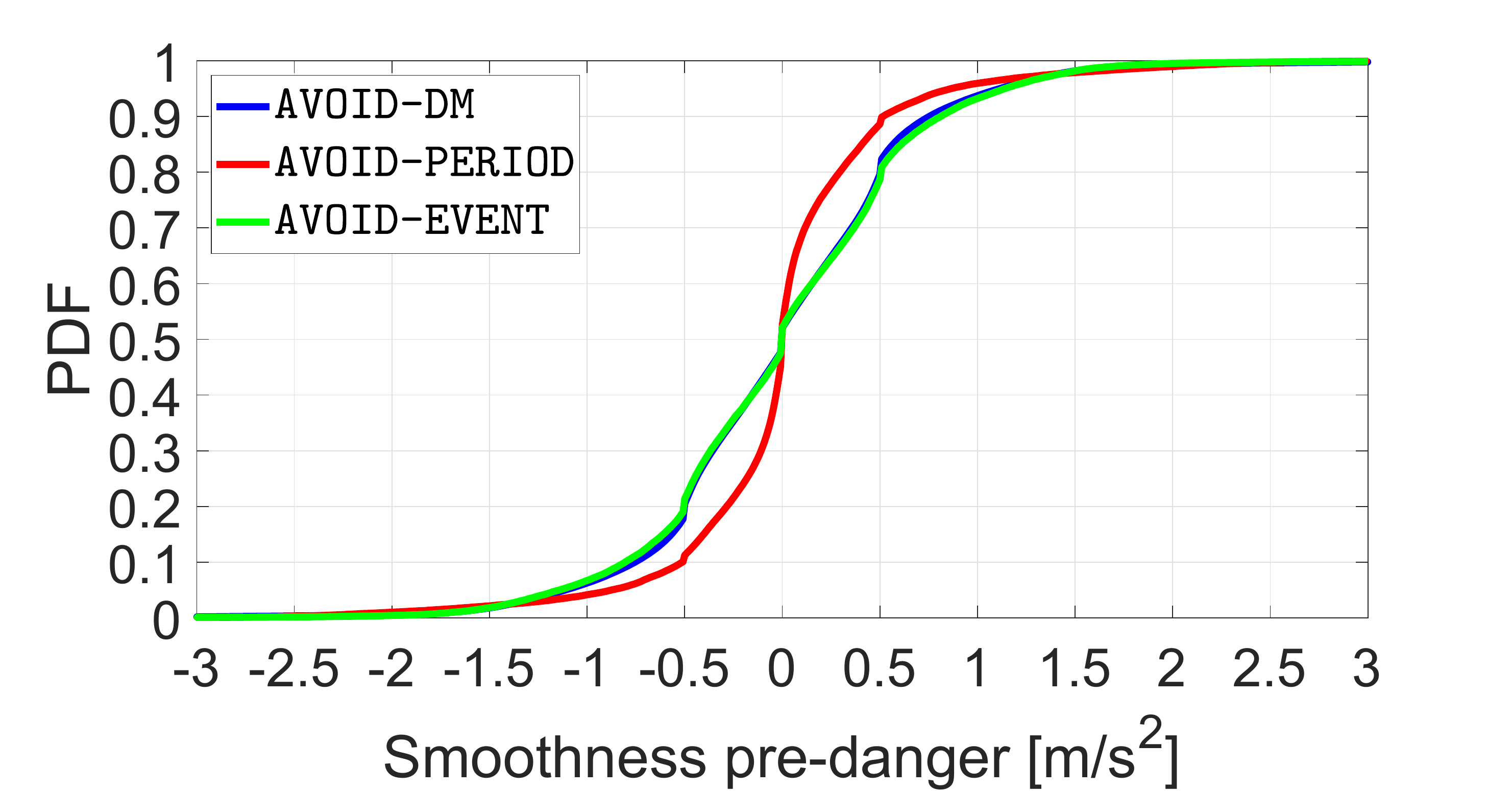}
  \label{fig:smoothness_predanger}
\end{subfigure}\\
\begin{subfigure}{.5\textwidth}
  \centering
  \includegraphics[width=0.8\columnwidth]{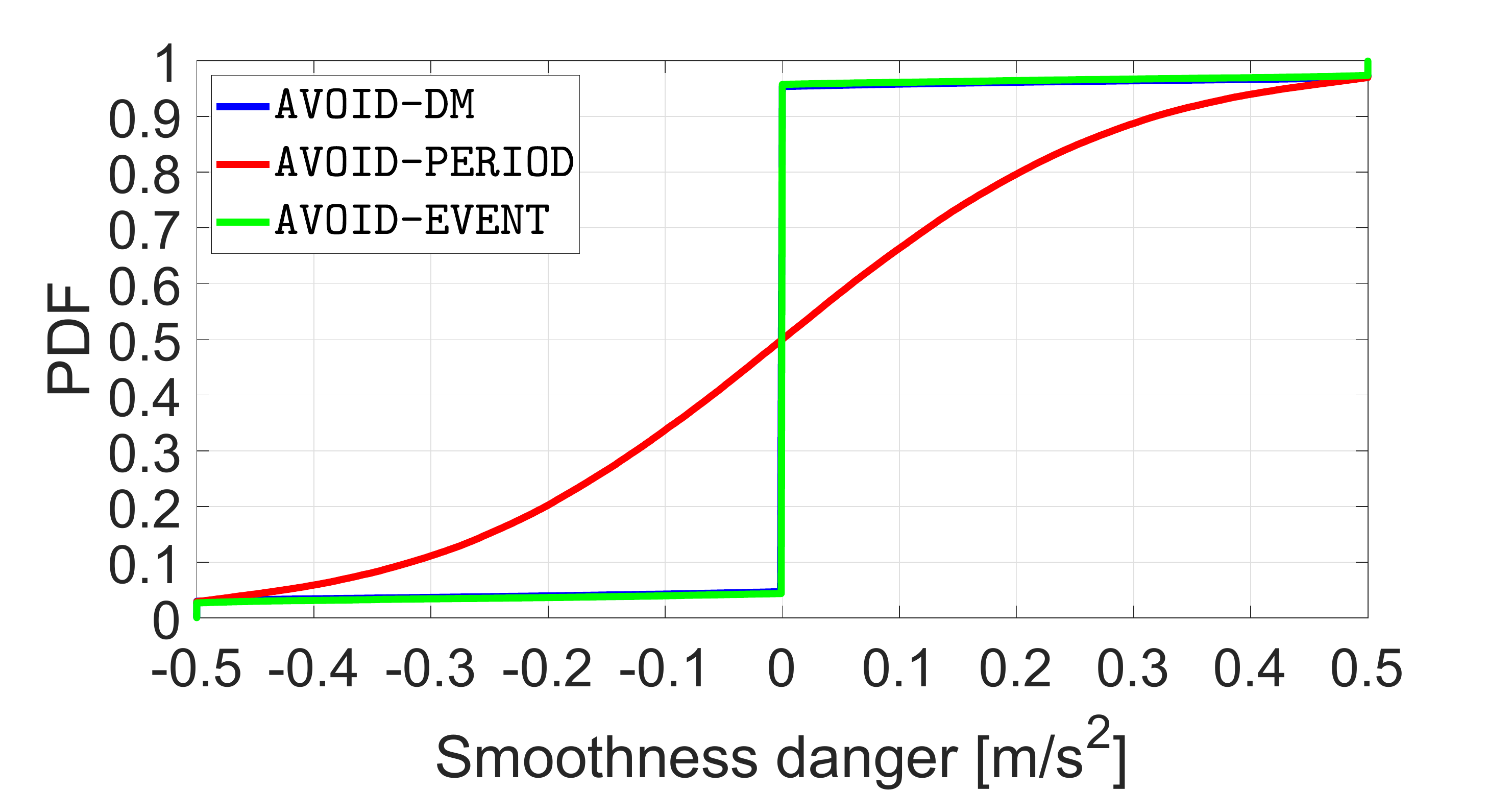}
  \label{fig:smoothness_danger}
\end{subfigure}
\caption{CAVs' consecutive acceleration controls in intersection.}
\label{fig:smoothness}
\end{figure}  

Finally, Fig. \ref{fig:smoothness} shows results on the acceleration applied by the CAVs for different areas of the intersection for both \texttt{AVOID-PERIOD} and \texttt{AVOID-EVENT} approaches. Specifically, in the pre-danger zone, \texttt{AVOID-PERIOD}, that recomputes the CAVs' controls at each slot, obtains less variable applied controls than \texttt{AVOID-EVENT} (Fig. \ref{fig:smoothness} (top)). This is due to the fact that \texttt{AVOID-EVENT}, when no event is triggered, tries to tightly track the safe trajectory decided by the IM, applying more diverse accelerations while coping for the deviations introduced by uncertainty. In the danger zone, \texttt{AVOID-EVENT} presents almost no control changes (Fig. \ref{fig:smoothness} (bottom)), as the CAVs, after choosing the appropriate time to enter the danger zone, proceed at constant speed (approximately $v_{MAX}=14$m/s). Instead, even though the intended trajectory is similar, \texttt{AVOID-PERIOD} ensures that the CAVs' speeds are exactly $v_{MAX}$ by re-planning at each time slot, hence, triggering continuous adjustments due to uncertainty. 

Clearly, smoother applied acceleration results in reduced gas consumption, when this is associated with a smaller average acceleration used (in absolute value) \cite{hadjigeorgiou2019optimizing}. Overall, accounting for both pre-danger and danger zones, \texttt{AVOID-EVENT} shows a $10.4\%$ reduction in gas consumption compared to \texttt{AVOID-PERIOD}, as the overall average absolute value of the difference among consecutive accelerations is equal to $0.227$ m/s$^2$ and $0.254$ m/s$^2$, respectively.

\section{CONCLUSIONS}
\label{sec:conclusion}
A novel framework for autonomous intersection management (\texttt{AVOID-PERIOD}) is presented, that considers uncertainty in the location of CAVs. This is a centralized approach that exploits periodic communication of update system state estimations and predictions between the CAVs and an IM, having as an aim the improvement of the performance as compared to existing techniques. This gain is mostly retained even when only a limited, intelligently chosen, number of control re-computations are executed, as in \texttt{AVOID-EVENT}. The event-triggering approach, not only improves consistently the intersection capacity, but it can easily scale for high-traffic scenarios. Future work includes considering CAVs turning, i.e., systems where the uncertainty may depend on the controls applied.

\bibliographystyle{IEEEtran}
\bibliography{IEEEabrv,roadsafety_ICCv2}

\end{document}